\documentclass[sigconf]{acmart}
\usepackage{booktabs}
\usepackage{tabularx}
\usepackage{array}
\usepackage{url}
\usepackage{booktabs}
\usepackage{graphicx}
\usepackage{makecell}
\usepackage{multirow}
\usepackage{pifont}

\newcommand{\cmark}{\ding{51}}
\newcommand{\xmark}{\ding{55}}
\usepackage{xspace}
\newcommand{\ourbench}{\textsc{{DuMateBench}}\xspace}
\AtBeginDocument{%
  }

\setcopyright{acmlicensed}
\copyrightyear{2018}
\acmYear{2027}
\acmDOI{XXXXXXX.XXXXXXX}
\acmConference[WSDM'27]{the 20th ACM International Conference on Web Search and Data Mining}{Feb 15--19,
  2027}{Hong Kong}
\acmISBN{978-1-4503-XXXX-X/2018/06}

\begin{document}


\title[DuMateBench: Evaluating Autonomous Agents in Complex Real-World Workflows]{DuMateBench: Evaluating Autonomous Agents in \\ Complex Real-World Workflows}

\settopmatter{authorsperrow=4}

\newcommand{\affRUC}{%
  \affiliation{%
    \institution{Renmin University of China}
    \city{Beijing}
    \country{China}}}

\newcommand{\affSDU}{%
  \affiliation{%
    \institution{Shandong University}
    \city{Jinan}
    \country{China}}}

\newcommand{\affIndependent}{%
  \affiliation{%
    \institution{Independent Researcher}
    \country{China}}}

\newcommand{\affMSU}{%
  \affiliation{%
    \institution{Michigan State University}
    \city{East Lansing}
    \country{USA}}}

\newcommand{\affNankai}{%
  \affiliation{%
    \institution{Nankai University}
    \city{Tianjin}
    \country{China}}}

\newcommand{\affBaidu}{%
  \affiliation{%
    \institution{Baidu, Inc.}
    \city{Beijing}
    \country{China}}}

\newcommand{\affECNU}{%
  \affiliation{%
    \institution{East China Normal University}
    \city{Shanghai}
    \country{China}}}

\newcommand{\affImperial}{%
  \affiliation{%
    \institution{Imperial College London}
    \city{London}
    \country{United Kingdom}}}

\author{Zechun Niu}
\authornote{Zechun Niu and Yukun Zhao contributed equally to this work.}
\affRUC
\email{niuzechun@ruc.edu.cn}

\author{Yukun Zhao}
\authornotemark[1]
\affSDU
\email{zhaoyukun@sdu.edu.cn}

\author{Jiaxin Zhang}
\affIndependent

\author{Xu Shen}
\author{Jinhua Si}
\affMSU

\author{Han Tian}
\affNankai

\author{Can Xu}
\affECNU

\author{Yunfan Song}
\affImperial

\author{Jiaxin Mao}
\affRUC

\author{Yansong Gao}
\author{Yuchen Li}
\author{Jianmin Wu}
\affBaidu


\author{Lingyong Yan}
\authornote{Co-corresponding authors.}
\affBaidu
\email{yanlingyong@baidu.com}

\author{Shuaiqiang Wang}
\authornotemark[2]
\affBaidu
\email{wangshuaiqiang@baidu.com}

\author{Dawei Yin}
\authornotemark[2]
\affBaidu
\email{yindawei@acm.org}

\renewcommand{\shortauthors}{Niu, Zhao et al.}

\begin{abstract}

Autonomous agents are increasingly adopted to complete complex, multi-tool workflows in real-world settings. 
However, existing benchmarks typically separate tasks by application or capability and evaluate agents in environments that are cleaner and more stable than those encountered in practice.
We introduce \ourbench, a real-session benchmark reconstructed from anonymized and privacy-screened user sessions collected from a large-scale production agent platform.
Each task preserves the relevant pre-solution interaction history, persistent configurations, and workspace state, and is then validated through human verification. 
The resulting benchmark comprises 200 tasks spanning 8 broad scenarios and 17 fine-grained capability categories, with most tasks requiring multiple capability coordination.
We execute these tasks in isolated Docker containers injected with three forms of real-world environmental complexity: \textsc{Insufficient}, \textsc{Unstable}, and \textsc{Noisy}, and assess performance using a hybrid deterministic and LLM-as-Judge evaluation protocol.
Experiments across five representative autonomous-agent frameworks paired with four state-of-the-art LLMs reveal substantial gaps in strict task completion. Complementary robustness, efficiency, and diagnostic analyses further show that performance under environmental perturbations is jointly shaped by the capabilities of the LLM and the surrounding agent framework.
The code and data are publicly available at \url{https://dumatebench.com/}.


\end{abstract}



\maketitle

\section{Introduction}
\label{sec:introduction}

\begin{table}[t]
\caption{Comparison with representative workflow-oriented and
environment-robustness benchmarks. A check mark indicates explicit
benchmark-level coverage rather than incidental occurrence in individual
tasks.}
\label{tab:benchmark-landscape}
\centering
\scriptsize
\setlength{\tabcolsep}{3.2pt}
\renewcommand{\arraystretch}{1.02}

\textbf{(a) Data source and execution environment}

\vspace{2pt}
\resizebox{\columnwidth}{!}{%
\begin{tabular}{@{}lccccc@{}}
\toprule
\textbf{Benchmark} &
\textbf{Year} &
\makecell{\textbf{Real-user}\\\textbf{Session}} &
\multicolumn{3}{c}{\textbf{Execution Environment}} \\
\cmidrule(lr){4-6}
& & &
\textbf{Insufficient} &
\textbf{Unstable} &
\textbf{Noisy} \\
\midrule

\multicolumn{6}{@{}l}{\textit{Multi-tool workflow benchmarks}} \\

OfficeBench~\cite{wang2024officebench}
& 2024 & \xmark
& \xmark & \xmark & \xmark \\

OdysseyBench~\cite{wang2025odysseybench}
& 2025 & \xmark
& \xmark & \xmark & \xmark \\

APEX-Agents~\cite{vidgen2026apexagents}
& 2026 & \cmark
& \xmark & \xmark & \cmark \\

WorkBuddy Bench~\cite{tencent2026workbuddy}
& 2026 & \cmark
& \xmark & \xmark & \xmark \\

\addlinespace[3pt]
\multicolumn{6}{@{}l}{\textit{Environment and robustness benchmarks}} \\

ToolSandbox~\cite{lu2024toolsandbox}
& 2024 & \xmark
& \xmark & \xmark & \cmark \\

SetupBench~\cite{arora2025setupbench}
& 2025 & \cmark
& \cmark & \xmark & \xmark \\

ComplexMCP~\cite{li2026complexmcp}
& 2026 & \xmark
& \xmark & \cmark & \cmark \\

\midrule
\textbf{\ourbench}
& \textbf{2026} & \cmark
& \cmark & \cmark & \cmark \\
\bottomrule
\end{tabular}%
}

\vspace{6pt}

\textbf{(b) Task capabilities}

\vspace{2pt}
\resizebox{\columnwidth}{!}{%
\begin{tabular}{@{}lccccc@{}}
\toprule
\textbf{Benchmark} &
\makecell{\textbf{Document}\\\textbf{Reading}} &
\makecell{\textbf{Document}\\\textbf{Editing}} &
\makecell{\textbf{File}\\\textbf{Organization}} &
\makecell{\textbf{Coding}\\\textbf{}} &
\makecell{\textbf{Web}\\\textbf{Retrieval}} \\
\midrule

\multicolumn{6}{@{}l}{\textit{Multi-tool workflow benchmarks}} \\

OfficeBench
& \cmark & \cmark & \xmark & \xmark & \xmark \\

OdysseyBench
& \cmark & \cmark & \xmark & \xmark & \xmark \\

APEX-Agents
& \cmark & \cmark & \cmark & \xmark & \xmark \\

WorkBuddy Bench
& \cmark & \cmark & \xmark & \cmark & \xmark \\

\addlinespace[3pt]
\multicolumn{6}{@{}l}{\textit{Environment and robustness benchmarks}} \\

ToolSandbox
& \xmark & \xmark & \xmark & \xmark & \xmark \\

SetupBench
& \xmark & \xmark & \xmark & \xmark & \xmark \\

ComplexMCP
& \cmark & \cmark & \cmark & \xmark & \xmark \\

\midrule
\textbf{\ourbench}
& \cmark & \cmark & \cmark & \cmark & \cmark \\
\bottomrule
\end{tabular}%
\vspace{-0.6cm}
}

\end{table}



Powered by large language models (LLMs), autonomous agents are increasingly deployed to tackle complex tasks across software engineering~\cite{xu2025agentcompany,glm5team2026glm5}, web-based knowledge work~\cite{boisvert2024workarenaplusplus,zheng2026ebench,kimiteam2026kimi}, office productivity~\cite{wang2024officebench,wang2025odysseybench,tencent2026workbuddy}, and multimodal content creation~\cite{dong2024dreamllm,kimiteam2026kimi}.
To assess whether these capabilities translate into practical utility for end users, agents must be evaluated on tasks grounded in real-world use. This underscores the need for benchmarks that measure whether agents can reliably complete workflows under realistic contextual and environmental conditions.

Recent agent benchmarks have expanded to cover multi-step workflows, extended interaction histories, and professional tasks, as summarized in Table~\ref{tab:benchmark-landscape}. 
For instance, OfficeBench evaluates common productivity applications, OdysseyBench introduces long interaction histories into office workflows, APEX-Agents targets long-horizon professional tasks, and WorkBuddyBench spans office and coding domains~\cite{wang2024officebench,wang2025odysseybench,
vidgen2026apexagents,tencent2026workbuddy}.
Despite this progress, a critical challenge persists--\textbf{\textit{Challenge 1: limited cross-capability workflow composition}}. 
Existing datasets typically group tasks by individual applications or predefined capabilities. Consequently, they provide limited coverage of workflows that integrate document processing, information retrieval, coding, content generation, or other cross-application operations.
Evaluation on these tasks therefore provides limited evidence of an agent's ability in real-world scenarios, as authentic user requests demand the orchestration of multiple capabilities within a single end-to-end workflow~\cite{meng2026clawmark,zhang2026clawbench,li2026agencybench}.

Robustness-oriented benchmarks such as ToolSandbox~\cite{lu2024toolsandbox}, SetupBench~\cite{arora2025setupbench}, and ComplexMCP~\cite{li2026complexmcp} examine the remaining conditions, as compared in Table~\ref{tab:benchmark-landscape}. They typically isolate the conditions from session-grounded workflows that require coordinated tool use and heterogeneous artifact production. 
This gives rise to \textbf{\textit{Challenge 2: insufficient environmental realism}}. In practice, agents may encounter unavailable tools, missing dependencies, resource constraints, intermittent networks, API failures, timeouts, and distracting or corrupted files. 
Furthermore, the above two challenges severely complicate reliable evaluation. 
Documents, spreadsheets, presentations, and images often admit multiple valid solutions, making exhaustive grading criteria difficult to define. Because reference answers are rarely available for tasks derived from real-world interactions, human evaluators may inadvertently reject valid alternative solutions or overlook substantive errors~\cite{dong2026misscore}. Ultimately, as summarized in Table~\ref{tab:benchmark-landscape}, existing benchmarks rarely address compositional workflows, environmental complexity, and heterogeneous task capabilities within a unified setting. 


To address these challenges, we introduce \ourbench, a
\emph{real-user session-derived benchmark} for evaluating autonomous agents
on cross-capability workflows that require coordinating multiple productivity
tools under realistic, imperfect environmental conditions. \ourbench derives
its tasks from anonymized, privacy-screened user sessions collected from
DuMate, a production agent platform. We preserve the user-visible interaction
context and persistent configurations, reconstructing the task instance and
workspace state to match the environment in which the request originally
occurred. Human verification then retains only tasks that are faithful,
well-specified, solvable, free of solution leakage, and independently
evaluable. The benchmark comprises 200 tasks covering eight broad scenarios
and 17 fine-grained task types, with most workflows spanning multiple
scenarios and task types. These tasks frequently integrate content generation
with coding, document manipulation, or Web retrieval, thereby coupling
compositional workflows with challenging execution conditions.

\begin{figure*}[h]
    \centering
    \includegraphics[width=0.98 \textwidth]{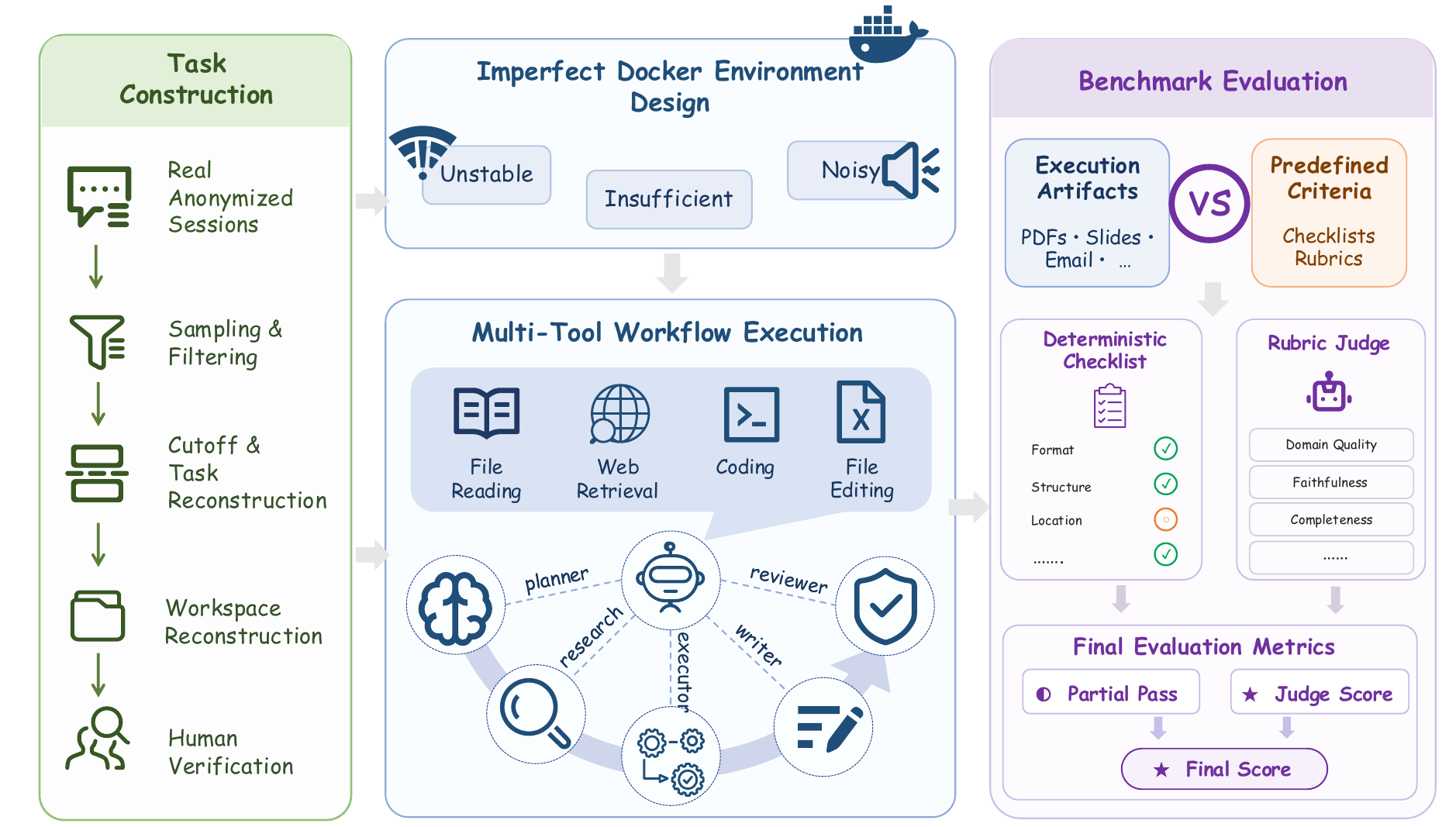}
    \caption{Overview of our proposed \ourbench. It is built through two stages: task construction and environment design, detailed in Sections~\ref{sec:task-construction} and~\ref{sec:environment-design}, respectively. Agents execute each task as a multi-tool workflow, and the resulting artifacts are assessed through two complementary channels: deterministic checklists for objectively verifiable requirements and a rubric-based LLM judge for semantic and presentation quality. The evaluation protocol is detailed in Section~\ref{sec:evaluation}.}
    \label{fig:overall-pipeline}
\end{figure*}

We design the \ourbench environment around three forms of real-world
complexity: \textsc{Insufficient} conditions with missing dependencies and
constrained resources, \textsc{Unstable} conditions with transient network and
tool failures, and \textsc{Noisy} conditions with distracting files and noisy
data. We execute each task in an isolated Docker container initialized with
its designed environment and reconstructed workspace. For evaluation, we
combine reviewed \emph{deterministic checklists} for explicit requirements
with artifact-specific \emph{LLM-as-Judge} rubrics for the correctness,
completeness, and quality of heterogeneous outputs. Using this protocol, we
evaluate five representative autonomous-agent frameworks paired with four
state-of-the-art base LLMs and conduct additional analyses of robustness to
workspace noise, efficiency, and failure modes.
Figure~\ref{fig:overall-pipeline} summarizes the task-construction,
environment-design, and evaluation pipeline of \ourbench.

Our contributions are summarized as follows:
\begin{itemize}
    \item \textbf{A real-session benchmark for compositional workflows.}
    We introduce \ourbench, comprising 200 executable tasks derived from real multi-turn DuMate sessions. By reconstructing the pre-task interaction history and workspace state, the benchmark preserves realistic context. It systematically evaluates workflows that integrate multiple capabilities across 8 high-level scenarios and 17 fine-grained task types.

    \item \textbf{Reproducible complex work environments.}
    We model three forms of real-world environmental complexity: \emph{insufficient} environments with missing tools, dependencies, or resources; \emph{unstable} environments with transient network and tool failures; and \emph{noisy} environments containing distracting files or noisy data. These conditions are instantiated
    in isolated Docker containers, enabling controlled and reproducible evaluation of agent reliability.

    \item \textbf{Comprehensive evaluation of autonomous agents.}
We evaluate 20 configurations of five autonomous-agent frameworks and four
LLMs on \ourbench, covering end-to-end performance, robustness to workspace
noise, efficiency, and failure modes. The results reveal strong agent--model
interactions, uneven robustness, and quality--efficiency trade-offs, while
trace analysis exposes weaknesses in execution planning, failure recovery,
and artifact verification. These findings show that \ourbench provides a
realistic and diagnostic evaluation of the capabilities required for complex
end-to-end workflows.

\end{itemize}

\section{Related Work}
\label{sec:related_work}

\subsection{Benchmarks for Multi-Tool Workflows}

The evaluation of LLM agents has increasingly shifted from isolated tool calls toward multi-step workflows that require the coordinated use of multiple tools~\cite{li2026clawsbench,zheng2026ebench,tencent2026workbuddy}.
For instance, OfficeBench~\cite{wang2024officebench} and SpreadsheetBench~\cite{ma2024spreadsheetbench} evaluate multi-step operations over office documents and spreadsheets, while WorkArena++~\cite{boisvert2024workarenaplusplus}, WorkBench~\cite{styles2024workbench}, and CRMArena~\cite{huang2024crmarena} extend evaluation to enterprise applications, structured data, and role-specific business processes. 
While OSWorld~\cite{xie2024osworld} evaluates cross-application tasks in a general computer environment and TheAgentCompany~\cite{xu2025agentcompany} embeds agents in a simulated software company, Workspace-Bench~\cite{tang2026workspacebench}, EnterpriseClawBench~\cite{zhong2026enterpriseclawbench}, and AgencyBench~\cite{li2026agencybench} emphasize file dependencies, workplace sessions, and extended real-world contexts. 
Their evaluation ranges from deterministic checks to rubrics, gold deliverables, and visual assessment~\cite{meng2026clawmark,vidgen2026apexagents,li2026agencybench,zhong2026enterpriseclawbench,wang2026finprobench}. 
Deterministic checks provide reproducible evidence for explicit requirements, while rubric-based LLM judges can assess the semantic, organizational, and perceptual quality of open-ended artifacts~\cite{peng2026ruverbench,dong2026misscore}.
Despite this progress, existing benchmarks isolate tasks within single applications, offering limited workflows that cut across different task types and tools. In reality, user requests frequently require the seamless integration of multiple capabilities.
\ourbench reconstructs multi-tool workflow tasks from anonymized real user sessions on a production agent platform that span multiple capabilities, including document processing, information retrieval, coding, content generation, and cross-application operations. This design enables grounded and reproducible evaluation of agents on authentic, complex user requests.


\vspace{-10pt}
\subsection{Benchmarks for Agent Reliability}

In real-user settings, agents rarely operate under ideal conditions. Recent work has therefore increasingly evaluated agent reliability under incomplete or unreliable execution conditions. 
ToolSandbox~\cite{lu2024toolsandbox} studies insufficient information
and distracting context, whereas EnvBench~\cite{eliseeva2025envbench} and SetupBench~\cite{arora2025setupbench} require agents to
resolve missing dependencies and construct incomplete software environments.
NIKA~\cite{wang2025nika} examines the diagnosis and recovery of dynamic network failures. Recent benchmarks extend reliability evaluation to a
broader range of perturbations: AgentNoiseBench~\cite{wang2026agentnoisebench} injects controllable user and
tool noise; ComplexMCP~\cite{li2026complexmcp} combines interdependent tools with unpredictable API
failures; OccuBench~\cite{hu2026occubench} introduces explicit errors and implicit data degradation;
and DeployBench~\cite{wang2026deploybench} covers incomplete or incompatible execution environments.
Together, these benchmarks assess information filtering, environment construction, fault diagnosis, and execution recovery.
However, these difficulties are generally isolated from session-derived workflows that require coordinated tool use and heterogeneous artifact
delivery. 
\ourbench instead integrates \emph{insufficient},
\emph{unstable}, and \emph{noisy} conditions into the same real-session tasks. 
This setting evaluates whether an agent can reliably fulfill user requests and deliver the required artifacts within realistic, imperfect environments.


\section{\ourbench Task Construction}
\label{sec:task-construction}

\ourbench is derived from anonymized, privacy-screened user sessions collected from a large-scale production agent platform~\footnote{https://www.dumate.cn/} serving millions of users. We first filter the collected sessions and reconstruct their task inputs, as described in Section~\ref{sec:task-derivation}. Human annotators then review each task and remove those with ambiguous intents, invalid test cases, or other critical issues (Section~\ref{sec:data-curation}). The overall task construction procedure is illustrated on the left panel of Figure~\ref{fig:overall-pipeline}. We present the task statistics in Section~\ref{sec:benchmark-statistics}.

\subsection{Task Derivation \& Reconstruction}
\label{sec:task-derivation}

The task derivation pipeline consists of three stages: interaction-history
reconstruction, cutoff-based instruction formulation, and workspace reconstruction.

\paragraph{Interaction-history reconstruction.} Our benchmark starts from anonymized, privacy-screened user sessions sampled from
a large-scale production agent platform. Each source session is represented by
a trace containing user messages, agent responses, tool interactions, system
events, and file operations. We restore the original event order and retain
the user-visible content, including user messages, displayed agent responses,
file references, and historical artifacts, while excluding internal execution
records such as tool calls, execution results, and orchestration messages. We
then discard sessions that are too simple to represent autonomous workflows,
including single-turn interactions and sessions with little tool or file
activity. The retained records form the ordered interaction history
$\mathcal{S}=(e_1,\ldots,e_n)$, where each $e_i$ denotes one retained
user-visible record, such as a user message, displayed agent response, file
reference, or historical artifact, ordered chronologically, and $n$ is the
total number of retained records.

\paragraph{Cutoff-based instruction formulation.}
For each retained interaction history, we use Claude Opus
4.8~\cite{anthropic2026opus48} to help select one user request as the target task.
A request may span multiple user turns when later turns refine the same
objective. We place the cutoff boundary $c$ immediately before the first user
turn of this target request. The user-visible events before $c$ form the
historical context
$\mathcal{H}_c=(e_1,\ldots,e_{c-1})$. We then consolidate all user turns
belonging to the target request into a self-contained task instruction $q_c$
that preserves the original objective, deliverables, file references, and
constraints. Agent responses, tool outputs, and generated artifacts produced
after the cutoff are excluded to prevent leakage from the original solution.


\paragraph{Workspace reconstruction.}
We use Claude Opus 4.8~\cite{anthropic2026opus48} to assist in reconstructing
the workspace state $\mathcal{W}_c$ available at the cutoff point. Given the
session trace and file-operation records, the model identifies the
user-uploaded files and historical agent artifacts available before $c$ and
checks their consistency with the reconstructed instruction and interaction
history. Post-cutoff files and workspace changes are excluded. If a
pre-existing file was overwritten, we restore its latest recoverable
pre-cutoff version; if an essential version cannot be recovered, we discard
the task. The resulting files are stored in \texttt{workspace\_seed/} and
copied to the working directory at runtime.

After these three stages, we obtain a task instance

\begin{equation}
    \mathcal{T}_c =
    \left(q_c,\, \mathcal{W}_c, \, \mathcal{M}_c\right),
\end{equation}

where $q_c$ denotes the task instruction,  $\mathcal{W}_c$ the reconstructed workspace, and $\mathcal{M}_c$
the task metadata and execution constraints. Each instance is serialized
into a standardized task package, while task-independent container and
environment components are materialized from shared infrastructure at
execution time.


\subsection{Human Verification}
\label{sec:data-curation}

We conduct human verification after candidate task reconstruction. Reviewers
inspect the source session, reconstructed interaction history, target
instruction, and workspace state. They assess whether
each task faithfully represents the original request, is solvable from the
provided state, contains no solution leakage, and is free of unresolved
privacy, security, or safety risks. A candidate task is retained only if it
satisfies all of the following criteria:

\begin{itemize}
\item \textbf{Request fidelity.}
The consolidated instruction preserves the intent, scope, deliverables, and
constraints of the original request. It introduces no requirements inferred
solely from the downstream agent response.

\item \textbf{Workspace completeness and consistency.}
The reconstructed workspace provide
sufficient and mutually consistent information and artifacts to complete the
task without irrecoverable ambiguity.

\item \textbf{Absence of solution leakage.}
The reconstructed task excludes all post-cutoff agent responses, tool results,
intermediate outputs, generated artifacts, and workspace changes that could
reveal the original solution.

\item \textbf{Privacy and security.}
Reviewers check for residual personally identifiable information, credentials,
API keys, private endpoints, confidential files, and other sensitive content.
They also inspect the task instructions, artifacts, and evaluation rules for
malicious, unsafe, or unauthorized operations. Tasks with unresolved risks are
removed or corrected without changing the original task semantics.

\end{itemize}

If a defect can be corrected without altering the original user request, we
reconstruct and re-review the task; otherwise, we discard it. We also exclude
any sample with unresolved ambiguity, inconsistent state, missing essential
context, solution leakage, residual privacy risks, or safety concerns. 

Retained tasks receive multi-label capability annotations under five coarse-grained scenarios: \emph{content generation} (text, image, video, and
audio generation); \emph{code development} (code writing and
generation); \emph{Web information retrieval} (Web information retrieval);
\emph{office document editing} (Word, spreadsheet, presentation, and PDF
creation or editing); \emph{office document reading} (Word, spreadsheet,
presentation, and PDF reading). These categories comprise
14 fine-grained capabilities, and each task may receive multiple labels.

\begin{figure}[t]
    \centering
    \includegraphics[
        width=0.52\textwidth
    ]{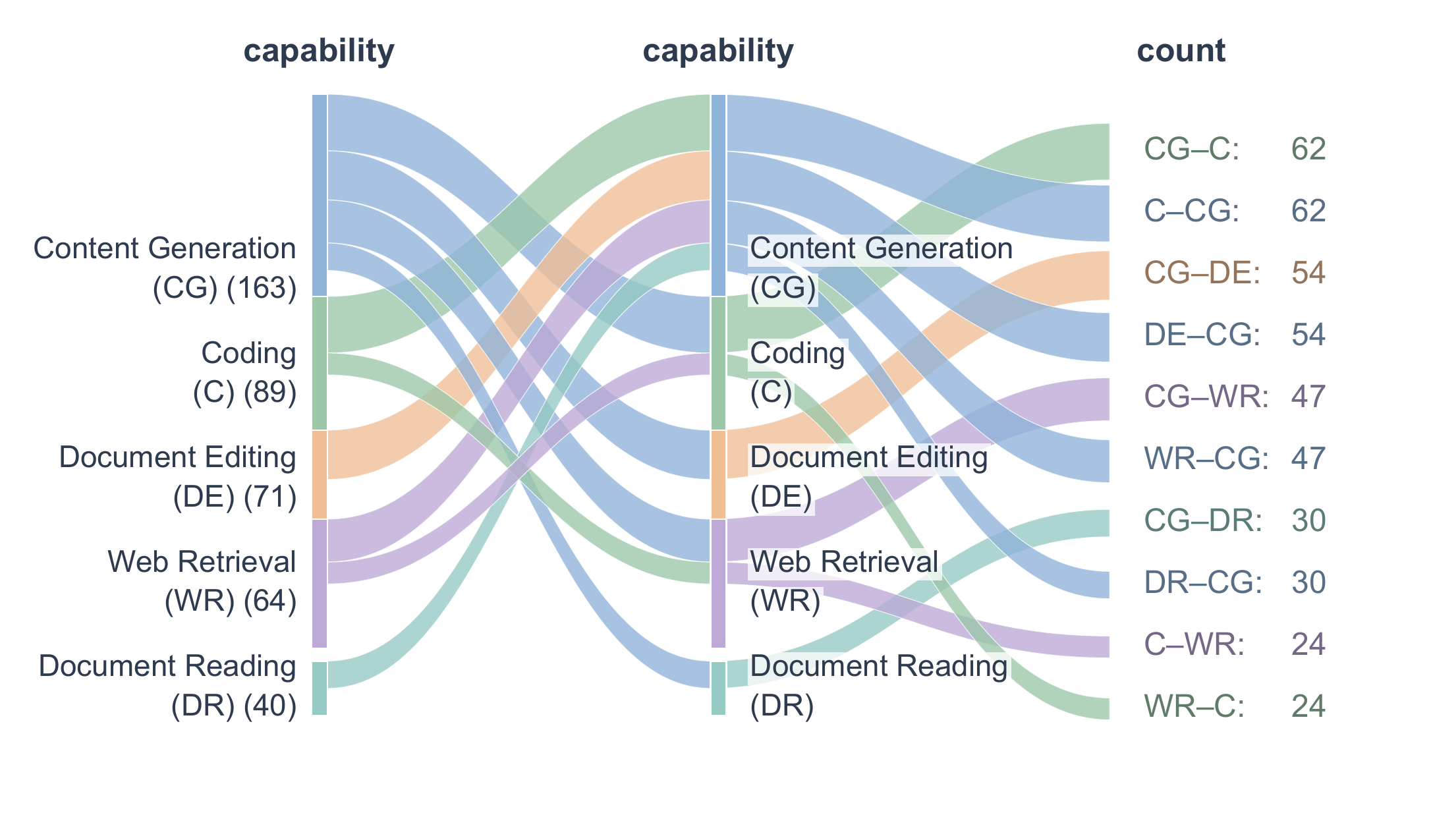}
    \caption{
        Coarse-grained scenario composition of \ourbench.
    }
    \label{fig:task_cross_category_overlap}
\end{figure}

\subsection{Benchmark Statistics}
\label{sec:benchmark-statistics}

We characterize \ourbench across three complementary dimensions: \textbf{capability coverage}, \textbf{capability compositionality}, and \textbf{knowledge-domain coverage}. The first two capture the required agent capabilities and the need to coordinate multiple capabilities within a single task, while the third captures subject domains.

\paragraph{Capability coverage.}
\ourbench contains 200 tasks annotated with 14 fine-grained capabilities
grouped into five coarse-grained scenarios. As shown in
Figure~\ref{fig:task_cross_category_overlap}, content generation is the most prevalent
scenario, followed by coding, document editing, Web information retrieval, and document reading (163, 89, 71, 64 and 40 respectively). At the fine-grained level, text generation and editing is the most frequent capability (148 tasks), followed by coding (89) and
information retrieval (64). The dataset contains 456 capability assignments in total,
averaging 2.28 capabilities per task. Since the annotations are multi-label,
the reported counts and percentages do not sum to 200 or $100\%$.

\paragraph{Capability compositionality.}
\ourbench is dominated by tasks that require coordinated capabilities: 159
tasks (79.50\%) span at least two coarse-grained scenarios, and 62 of these span
three or more. 
As shown in
Figure~\ref{fig:task_cross_category_overlap}, the most common cross-scenario combinations are code
development with content generation (62 tasks), office document editing with
content generation (54), and Web information retrieval with content generation
(47). These results indicate that the benchmark targets realistic
multi-capability workflows rather than isolated tool execution. 


\begin{figure}[t]
    \centering
    \includegraphics[width= 1.0\linewidth]{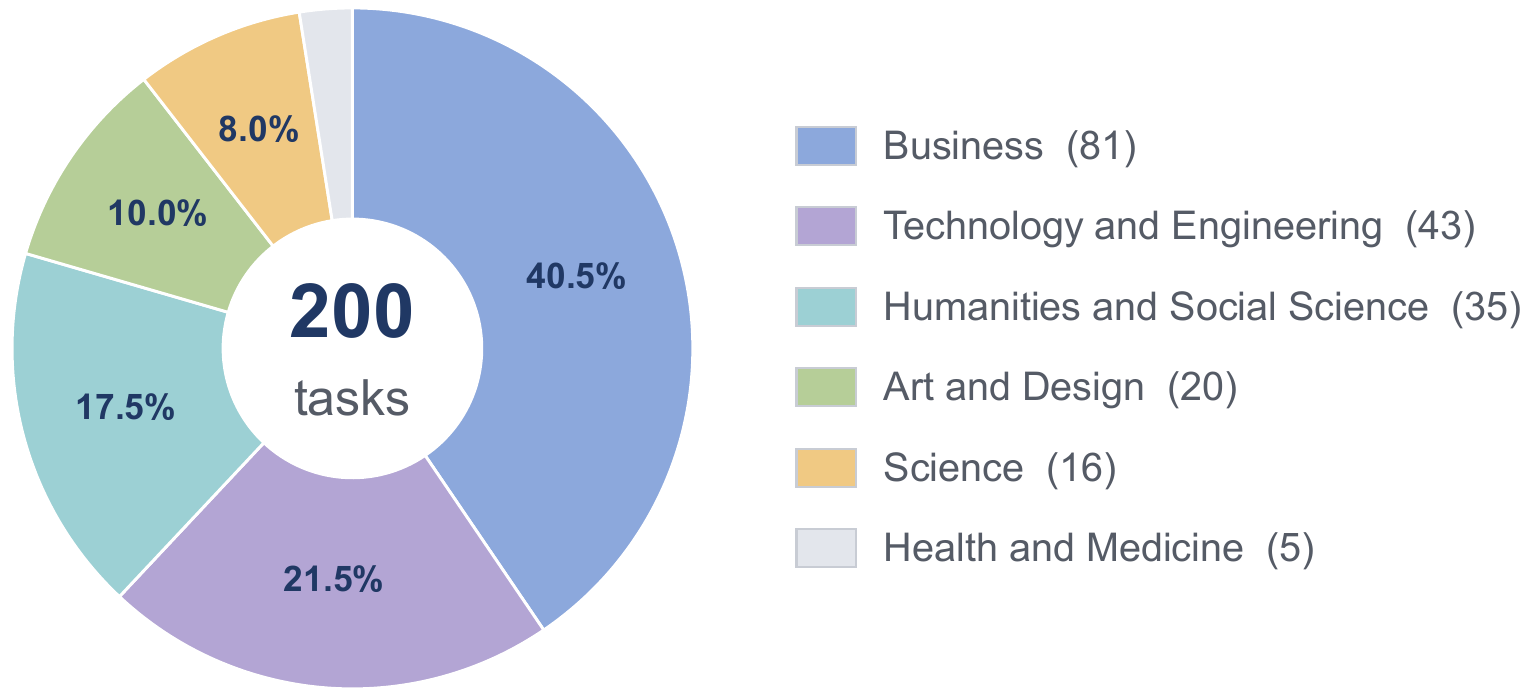}
    \caption{Knowledge-domain distribution of the 200 tasks included in the
 MMMU-derived topic analysis.}
    \label{fig:knowledge-domain-distribution}
\end{figure}

\paragraph{Knowledge-domain coverage.}
The capability taxonomy describes how a task is performed, but not the
knowledge required to complete it. We therefore assign each classified task a
single knowledge domain using an MMMU-derived taxonomy~\cite{yue2024mmmu}. The assignment follows
the core subject matter of the instruction rather than its delivery format. For
example, producing a financial report is categorized by its financial content
rather than as document editing. As shown in
Figure~\ref{fig:knowledge-domain-distribution}, the classified tasks span six
broad domains. Business is the largest category (40.5\%), followed by
technology and engineering (21.5\%) and humanities and social science
(17.5\%). At the finer level, the tasks cover 23 populated subjects, among
which computer science, finance, marketing, sociology, and management appear
most frequently. \ourbench therefore evaluates agents across varied
subject matter rather than within a single professional domain.

\section{\ourbench Environment Design}
\label{sec:environment-design}

Real-world autonomous agents rarely operate in pristine, fully prepared environments. Required tools may be missing, external services may fail mid-execution, and workspaces often contain irrelevant or conflicting data. We categorize these recurring environmental challenges into three dimensions: \textsc{Insufficient} (missing dependencies and constrained resources), \textsc{Unstable} (transient network and tool failures), and \textsc{Noisy} (distractor files and data). To model these conditions while retaining reproducibility, we execute each task in an isolated Docker container initialized with its reconstructed workspace. The evaluated agent runs as a non-root user and is confined to explicitly defined workspace, output, and logging boundaries.

\subsection{Insufficient Environment Design}
\label{sec:insufficient-environments}

\textsc{Insufficient} environments model missing dependencies and constrained resources. The base image provides general-purpose shell, networking, archive, and document utilities, but it does not guarantee that every task-specific system package, Python library, or office tool is preinstalled. Agents therefore need to inspect the environment, diagnose missing dependencies, install permitted packages when appropriate, or use alternative implementations.

Tasks may also impose explicit limits on CPU, memory, storage, and execution time. These constraints evaluate environment diagnosis, configuration, and resource-aware execution, rather than rewarding access to a fully prepared software stack. Evaluation runs outside the task container, so missing agent-side dependencies do not affect the evaluator.

\subsection{Unstable Environment Design}
\label{sec:unstable-environments}

\textsc{Unstable} environments model transient network and tool-execution failures. We inject faults at the network and tool layers under two schedules: startup faults create a reproducible initial failure window, while periodic faults introduce intermittent disruptions during longer executions. Network perturbations include DNS failures, IP/port blocking, and injected latency or packet loss. Besides, tool wrappers can produce temporary unavailability, delayed responses, missing fields in generated artifacts, and seeded nondeterministic timeouts.

Each fault configuration specifies its type, activation schedule, duration, probability, and random seed. Network controls apply only to agent-issued traffic; model-inference traffic is routed separately and remains exempt. Fault selection, activation, and recovery are recorded in structured logs. To complete a task, agents may need to distinguish transient failures and use retries, backoff, fallback tools, or alternative execution plans as appropriate.

\subsection{Noisy Environment Design}
\label{sec:noisy-environments}

\textsc{Noisy} environments model distractor files and data within the reconstructed workspace. We preserve noise already present in the original session data, including irrelevant, redundant, or outdated materials and similarly named files with different contents. To introduce controlled and reproducible variation, we use a noise generator to add task-specific distractor files and data, such as stale intermediate outputs, duplicate records, and irrelevant documents, without altering the inputs required to solve the task. This setting requires agents to identify relevant files, verify data provenance, and distinguish task-relevant evidence from distractors before acting.

In summary, building upon the task and workspace construction described in Section~\ref{sec:task-construction}, we instantiate each task in an isolated Docker container and inject the insufficient, unstable, and noisy conditions.
These designs enable reproducible evaluation of whether an agent can maintain reliable task completion in a complex and challenging environment. The environment is thus an explicit, controlled, and auditable component of benchmark difficulty rather than merely an execution container.
\section{\ourbench Evaluation}
\label{sec:evaluation}

\ourbench evaluates explicit requirement satisfaction and overall artifact quality through two complementary mechanisms. Deterministic checklist evaluation verifies objectively testable constraints. Rubric-based evaluation utilizes an artifact-specific LLM judge to assess semantic, organizational, and perceptual properties that cannot be captured by fixed rules. The overall evaluation protocol is illustrated on the right panel of Figure~\ref{fig:overall-pipeline}.

\subsection{Deterministic Checklist Evaluation}
\label{sec:checklist-evaluation}

Each task is associated with a set of atomic checks generated by an LLM and
subsequently reviewed by human annotators. The checks cover output existence
and location, format validity, required or forbidden content, document
structure, spreadsheet values and formulas, and the integrity of protected
files. For Web-retrieval tasks, we curate evaluator-only reference sources;
for numerical and question-answering tasks with objectively verifiable
targets, we provide human-checked gold answers. None of these evaluation
materials is exposed to the agent.

For task $t$, let $c_{t,i}\in\{0,1\}$ denote whether check $i$ is satisfied
and $n_t$ the number of checks. The task-level partial pass rate is
\begin{equation}
    P_t=\frac{1}{n_t}\sum_{i=1}^{n_t}c_{t,i}.
\end{equation}
It measures the proportion of atomic requirements satisfied, thereby
crediting partial progress. A task is considered complete only when all of its
deterministic checks pass.

\paragraph{Checklist coverage statistics.}
After applying the eight-task exclusion used in the final evaluation, the
inventory contains 200 tasks and 1,257 atomic checks, averaging 6.29 checks per
task. Existence checks account for 482 items (38.35\%), required-content
checks for 459 (36.52\%), and format-validity checks for 228 (18.14\%).
Together, these categories constitute 93.01\% of all checks. The remainder
comprises directory-structure checks (34, 2.70\%), forbidden-content checks
(21, 1.67\%), and specialized evaluators (33, 2.62\%). Thus, the checklist
primarily verifies artifact presence, validity, and compliance with explicit
requirements before qualitative evaluation.

\subsection{Rubric-Based Artifact Evaluation}
\label{sec:rubric-evaluation}

Deterministic checks alone cannot establish whether a report adequately
addresses its audience, a workbook provides a useful analysis, or a
presentation is visually coherent. \ourbench therefore uses
artifact-specific judges for textual documents, presentations, spreadsheets,
PDFs, images, audio, and video. The judge receives the task instruction,
candidate artifact, and relevant reference material in modality-appropriate
forms, such as extracted text, document structure, spreadsheet formulas,
rendered pages, images, or sampled video frames.

For each task, an LLM generates an initial rubric from the instruction,
deterministic checks, and reference inventory before any system output is
evaluated. Human annotators then review its clarity, coverage, and
assessability. The resulting rubric and reference set are fixed and shared
across all agent--model configurations, preventing the evaluation criteria
from adapting to a particular output.

Each rubric contains between three and sixteen atomic criteria with normalized
weights and anchored score levels from 0 to 4. The judge assigns a score and
supporting evidence to each criterion. A criterion is marked
\texttt{cannot\_assess} when the artifact does not provide sufficient evidence.
\paragraph{Rubric coverage statistics.}
The filtered inventory contains 454 artifact-specific criteria JSON files
covering 197 of the 200 tasks. These files define 2,308 atomic criteria,
corresponding to 11.54 criteria per task and 5.08 criteria per rubric file on
average. The observed rubric files contain three to eleven criteria, within the
design range above. By artifact suffix, Markdown files are most frequent (108
files, 23.79\%), followed by SVG (60, 13.22\%), PNG (51, 11.23\%), DOCX (45,
9.91\%), Python (43, 9.47\%), and HTML (34, 7.49\%); these six types comprise
75.11\% of rubric files. Across all criteria, the dominant dimensions are
requirement completeness (790 criteria, 34.23\%), presentation readability
(275, 11.92\%), functional correctness (251, 10.88\%), content relevance
(245, 10.62\%), and factual correctness and faithfulness (149, 6.46\%).
The five dimensions together account for 67.63\% of criterion instances,
while the remaining dimensions provide modality- and artifact-specific
coverage such as visual hierarchy, edge-case robustness, and reference
fidelity.

Let $\alpha_{t,a,j}$ be the normalized weight of criterion $j$ for artifact
$a$, and let $s_{t,a,j}\in\{0,1,2,3,4\}$ denote its score. The artifact-level
judge score is
\begin{equation}
    J_{t,a}
    =
    \sum_j \alpha_{t,a,j}
    \frac{\tilde{s}_{t,a,j}}{4},
    \qquad
    \tilde{s}_{t,a,j}
    =
    \begin{cases}
        s_{t,a,j}, & \text{if assessed},\\
        0,         & \text{otherwise}.
    \end{cases}
\end{equation}
Unassessed criteria receive zero contribution, preventing incomplete evidence
from increasing the score. When repeated judge runs are used, criterion-level
scores are aggregated by their median.

\subsection{Score Aggregation}
\label{sec:evaluation-aggregation}

Let $\mathcal{A}_t$ denote the set of supported target artifacts for task $t$.
The task-level judge score is the macro-average
\begin{equation}
    J_t =
    \frac{1}{|\mathcal{A}_t|}
    \sum_{a\in\mathcal{A}_t}J_{t,a}.
\end{equation}
A missing expected artifact receives a score of zero. Unsupported artifact
types are recorded but omitted from the artifact average. For a task to which
no artifact-specific judge applies by design, the deterministic score is used
as its task score.

The final score gives 30\% weight to deterministic requirement coverage and
70\% weight to artifact quality:
\begin{equation}
    F_t = 0.3P_t + 0.7J_t.
\end{equation}
We report $P_t$, $J_t$, and $F_t$ separately and macro-average each
metric across tasks.


\section{Experiments}
\label{sec:experiments}

In this section, we evaluate autonomous agents across four complementary dimensions on \ourbench to answer the following research questions (RQs):
\begin{itemize}
    \item \textbf{RQ1.} How do autonomous agents perform on \ourbench?
    \item \textbf{RQ2.} How does environmental noise affect the performance of autonomous agents on \ourbench?
    \item \textbf{RQ3.} How efficiently do autonomous agents solve tasks on \ourbench?
    \item \textbf{RQ4.} What are the common failure modes of autonomous agents on \ourbench?
\end{itemize}

\subsection{Experimental Settings}
\label{sec:experimental-settings}

\paragraph{Agents and models.}
We evaluate five representative autonomous agents: Claude Code
(v2.1.212)~\cite{anthropic2026claudecode}, Hermes (v0.19.0)~\cite{nousresearch2026hermes},
DuMate (v1.0.59)~\cite{yan2026dumate}, OpenCode (v1.18.4)~\cite{opencode2026},
and OpenClaw (v2026.7.1-2)~\cite{openclaw2026}. Each agent is paired with four
base models: GPT-5.5~\cite{openai2026gpt55}, Opus-4.8~\cite{anthropic2026opus48},
GLM-5.2~\cite{glm5team2026glm5}, and DeepSeek-V4-Pro~\cite{deepseek2026v4}, yielding 20 agent--model configurations. For each configuration, we preserve the
agent’s native control loop, tool-use policy, and model-interface protocol. We retain runtime-specified settings and do
not standardize decoding parameters across runtimes. 

\paragraph{Execution Harness.}
For each task, all agent configurations receive the same instruction and
initial workspace. Each trial is executed by a non-root agent in an isolated
Docker container with a fresh workspace at \texttt{/workspace}. Evaluation
files and references remain inaccessible to the agent. After execution or
timeout, the final workspace state is preserved and evaluated.

\paragraph{Environment Settings.}
Following the three environment challenges defined in
Section~\ref{sec:environment-design}, we evaluate agents under
\textsc{Insufficient}, \textsc{Unstable}, and \textsc{Noisy} conditions. First,
the \textsc{Insufficient} condition models missing capabilities and limited
resources in the real world. Each task runs in an isolated Docker container based on
\texttt{python:3.12-slim}. The image provides only general-purpose shell,
archive, networking, PDF, and process utilities, including \texttt{bash},
\texttt{curl}, \texttt{git}, \texttt{dnsutils}, \texttt{iproute2},
\texttt{iptables}, \texttt{jq}, \texttt{poppler-utils}, \texttt{procps},
\texttt{unzip}, and \texttt{vim-tiny}. The task-specific system tools and Python
packages are not preinstalled. Besides,
each container is limited to 2 CPUs, 8\,GB of memory, 12\,GB of storage, and a
wall-clock budget of 1,800 seconds.

Second, the \textsc{Unstable} condition introduces controlled failures at the
network and tool layers. At startup, DNS failure, latency with packet loss, and
destination blocking are each enabled for 8 seconds. During execution, a fault
daemon independently samples these three network faults every 45 seconds with
probabilities of 0.35, 0.45, and 0.25, respectively. When selected, DNS
failure, latency with packet loss, and destination blocking remain active for
6, 10, and 8 seconds, respectively. In addition, The tool layer models transient OCR unreliability: it forces the first eligible OCR call to fail and, with probability 0.4, delays a subsequent OCR response by 5 seconds.

Third, the \textsc{Noisy} condition retains natural noise from the reconstructed
workspace, including historical files, temporary notes, and artifacts
unrelated to the current task. For RQ2, we additionally inject seeded
synthetic distractors, such as similarly named, outdated, duplicate, or
conflicting files, to control noise intensity.

\paragraph{Evaluation.}
For each run, we apply the evaluation protocol described in
Section~\ref{sec:evaluation}. After the agent finishes, a Python evaluator first executes the task's deterministic checklist and computes the partial pass rate $P$, the fraction of checklist requirements that are satisfied. 
We then utilize Gemini-3.1-Pro-Preview~\cite{google2026gemini31pro} as a judge model to evaluate the quality of the outputted artifacts. The judge
receives the task instruction, candidate artifacts, and
task-relevant reference files, and scores each artifact with predefined task-specific rubrics. We report the partial
pass rate $P$, average artifact-judge score $J$ together with the combined final score $F=0.3P+0.7J$. In addition, we measure each run's wall-clock time and token usage, including input, output, and total tokens, to characterize efficiency when answering RQ3.

\subsection{RQ1. How Do Autonomous Agents Perform on \ourbench?}
\label{sec:main-results}

\begin{table*}[htbp]
    \caption{Results of autonomous agents on 200 \ourbench tasks. ``Partial'' denotes the partial pass rate, ``Judge'' denotes the LLM judge score, and ``Final'' denotes the final score (computed as
    $0.3\,\mathrm{Partial}+0.7\,\mathrm{Judge}$). The best value for each metric within each model block is shown in bold.}
    \label{tab:main-results}
    \centering
    \scriptsize
    \resizebox{\linewidth}{!}{%
    \begin{tabular}{lrrrrrrrrrrrr}
        \toprule
        \textbf{Agent}
        & \multicolumn{3}{c}{\textbf{GPT-5.5}}
        & \multicolumn{3}{c}{\textbf{Opus-4.8}}
        & \multicolumn{3}{c}{\textbf{GLM-5.2}}
        & \multicolumn{3}{c}{\textbf{DeepSeek-V4-Pro}} \\
        \cmidrule(lr){2-4}
        \cmidrule(lr){5-7}
        \cmidrule(lr){8-10}
        \cmidrule(lr){11-13}
        & \textbf{Partial} & \textbf{Judge} & \textbf{Final}
        & \textbf{Partial} & \textbf{Judge} & \textbf{Final}
        & \textbf{Partial} & \textbf{Judge} & \textbf{Final}
        & \textbf{Partial} & \textbf{Judge} & \textbf{Final} \\
        \midrule
        Claude Code
        & 0.8613 & 0.7494 & 0.7830
        & 0.8734 & 0.7884 & 0.8139
        & 0.7317 & 0.6398 & 0.6674
        & 0.8373 & 0.7915 & 0.8052 \\

        Hermes
        & 0.9001 & 0.7634 & 0.8044
        & 0.8986 & 0.7836 & 0.8181
        & 0.8253 & 0.7213 & 0.7525
        & 0.8521 & 0.8096 & 0.8223 \\

        DuMate
        & \textbf{0.9025} & \textbf{0.7768} & \textbf{0.8145}
        & \textbf{0.9088} & \textbf{0.8316} & \textbf{0.8548}
        & \textbf{0.8829} & \textbf{0.7711} & \textbf{0.8046}
        & 0.8631 & \textbf{0.8229} & \textbf{0.8350} \\

        OpenCode
        & 0.7696 & 0.6568 & 0.6906
        & 0.8555 & 0.7736 & 0.7982
        & 0.8613 & 0.7396 & 0.7761
        & \textbf{0.8721} & 0.7716 & 0.8017 \\

        OpenClaw
        & 0.8279 & 0.7590 & 0.7797
        & 0.6127 & 0.5690 & 0.5821
        & 0.7668 & 0.7075 & 0.7253
        & 0.8090 & 0.7800 & 0.7887 \\
        \bottomrule
    \end{tabular}%
    }

\end{table*}

\paragraph{Sensitivity across agent systems.}
Table~\ref{tab:main-results} shows that sensitivity to the base model varies
across agent systems. DuMate has the smallest Final-score range, spanning
0.8046--0.8548 (5.02 percentage points), followed by Hermes at
0.7525--0.8223 (6.98 points). OpenCode varies from 0.6906 to 0.8017
(11.11 points), and Claude Code from 0.6674 to 0.8139 (14.65 points).
OpenClaw is the most sensitive, with scores ranging from 0.5821 to 0.7887
(20.66 points). The best-performing model also differs by agent: Opus-4.8
performs best with Claude Code and DuMate, whereas DeepSeek-V4-Pro performs
best with Hermes, OpenCode, and OpenClaw. DuMate ranks first within each model
block, but its margin over the strongest alternative varies from 1.01 points
under GPT-5.5 to 3.67 points under Opus-4.8, with intermediate margins of 2.85
and 1.27 points under GLM-5.2 and DeepSeek-V4-Pro.
\paragraph{Compatibility across base models.}
The base models likewise differ in their consistency across agent systems.
DeepSeek-V4-Pro achieves the highest mean Final score (0.8106), followed by
GPT-5.5 (0.7744), Opus-4.8 (0.7734), and GLM-5.2 (0.7452).
DeepSeek-V4-Pro also has the smallest cross-agent range, with scores between
0.7887 and 0.8350 (4.63 percentage points). GPT-5.5 and GLM-5.2 show wider
ranges of 12.39 and 13.72 points, respectively. Although Opus-4.8 produces the
highest individual score in the table (0.8548), its scores span
0.5821--0.8548, yielding the largest cross-agent range of 27.27 points. Thus,
GPT-5.5 and Opus-4.8 have nearly identical mean scores, differing by only
0.0010, despite markedly different variation across agent systems.
\subsection{RQ2. How Does Environmental Noise Affect Agent Performance?}
\label{sec:noise-results}

Real-world workspaces often contain stale, duplicated, or irrelevant files.
We evaluate five agents with Opus-4.8 under four noise levels: normal, low,
medium, and high. The normal condition retains only the natural noise present
in the reconstructed workspace, as in RQ1. The other conditions introduce
seeded filename and content distractors, including backups, historical
versions, duplicated text, conflicting values, and corrupted content. Low,
medium, and high noise perturb approximately 35\%, 65\%, and 100\% of
workspace files, capped at three, six, and ten files per task, respectively.
Higher levels also increase the number of distractors generated for each
selected file.

\begin{figure}[t]
    \centering
    \includegraphics[width=\linewidth]
        {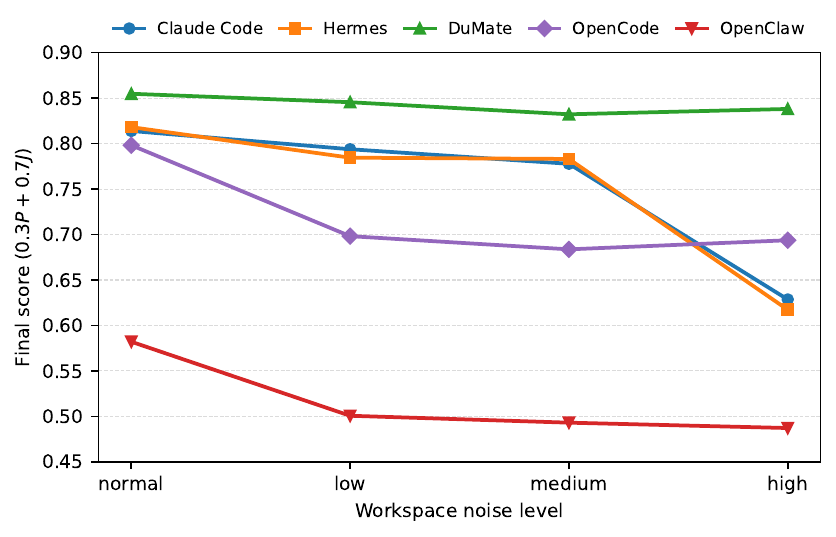}
    \caption{Final scores of five agents with Opus-4.8 under increasing
    workspace noise.}
    \label{fig:opus-noise-performance}
\end{figure}

As shown in Figure~\ref{fig:opus-noise-performance}, performance generally
declines as workspace noise increases. From the normal to the high-noise
condition, Hermes, Claude Code, OpenCode, and OpenClaw lose 20.08, 18.53,
10.45, and 9.49 percentage points, respectively. DuMate decreases from 0.8548
to 0.8381, a loss of 1.67 percentage points. Under this experimental setting,
the magnitude of degradation therefore varies considerably across agent
systems, with DuMate exhibiting the smallest decrease among the evaluated
agents.

\subsection{RQ3. How Efficiently Do Autonomous Agents Solve Tasks on \ourbench?}
\label{sec:efficiency-results}

\begin{table}[t]
    \centering
    \caption{Efficiency results for the 20 agent--model configurations on \ourbench. All quantities are averaged per
    task. DS-V4-Pro denotes DeepSeek-V4-Pro for short.}
    \label{tab:efficiency}
    \footnotesize
    \setlength{\tabcolsep}{1.5pt}
    \resizebox{\columnwidth}{!}{%
    \begin{tabular}{llrrrrr}
        \toprule
        \textbf{Agent} & \textbf{Model} & \textbf{Final score} & \textbf{Time (s)} & \textbf{Input tokens} & \textbf{Output tokens} & \textbf{Total tokens} \\
        \midrule
        Claude Code & GPT-5.5 & 0.7830 & \textbf{274.99} & 281,539 & 7,625 & 289,164 \\
        Claude Code & Opus-4.8 & 0.8139 & 464.43 & 336,549 & 14,744 & 351,293 \\
        Claude Code & GLM-5.2 & 0.6674 & 943.78 & 4,579,662 & 24,268 & 4,603,930 \\
        Claude Code & DS-V4-Pro & 0.8052 & 784.96 & 3,043,966 & 24,527 & 3,068,493 \\
        \midrule
        Hermes & GPT-5.5 & 0.8044 & 376.43 & 356,906 & 9,608 & 366,514 \\
        Hermes & Opus-4.8 & 0.8181 & 517.24 & 757,359 & 14,032 & 771,391 \\
        Hermes & GLM-5.2 & 0.7525 & 810.01 & 1,059,625 & 20,444 & 1,080,069 \\
        Hermes & DS-V4-Pro & 0.8223 & 847.90 & 1,151,063 & 29,407 & 1,180,470 \\
        \midrule
        DuMate & GPT-5.5 & 0.8145 & 518.00 & 868,023 & 8,636 & 876,659 \\
        DuMate & Opus-4.8 & \textbf{0.8548} & 1038.74 & 1,548,844 & 13,280 & 1,562,124 \\
        DuMate & GLM-5.2 & 0.8046 & 694.91 & 2,231,592 & 23,339 & 2,254,931 \\
        DuMate & DS-V4-Pro & 0.8350 & 821.65 & 1,730,007 & 36,219 & 1,766,226 \\
        \midrule
        OpenCode & GPT-5.5 & 0.6906 & 368.88 & \textbf{269,280} & \textbf{4,252} & \textbf{273,532} \\
        OpenCode & Opus-4.8 & 0.7982 & 603.72 & 762,894 & 13,216 & 776,110 \\
        OpenCode & GLM-5.2 & 0.7761 & 448.20 & 1,191,097 & 13,049 & 1,204,146 \\
        OpenCode & DS-V4-Pro & 0.8017 & 513.37 & 1,029,007 & 9,079 & 1,038,086 \\
        \midrule
        OpenClaw & GPT-5.5 & 0.7797 & 293.42 & 485,955 & 8,239 & 494,194 \\
        OpenClaw & Opus-4.8 & 0.5821 & 300.12 & 809,440 & 7,988 & 817,428 \\
        OpenClaw & GLM-5.2 & 0.7253 & 389.75 & 908,673 & 12,847 & 921,520 \\
        OpenClaw & DS-V4-Pro & 0.7887 & 482.55 & 1,215,111 & 20,051 & 1,235,162 \\
        \bottomrule
    \end{tabular}%
    }
\end{table}

We characterize
efficiency along two dimensions, latency and token use, rather than treating it
as a single scalar quantity. We report the efficiency for each agent--model configuration on \ourbench in table~\ref{tab:efficiency}, where input token denotes the aggregation of standard input tokens and cache-read tokens and all values are means per task.

The results reveal a clear quality--efficiency trade-off. DuMate with Opus-4.8
achieves the highest Final score (0.8548), but it is also the slowest
configuration (1,038.74~s per task) and consumes 1.56M total tokens per task.
At the other end of the latency spectrum, Claude Code with GPT-5.5 is the
fastest configuration (274.99~s) and achieves a Final score of 0.7830. OpenClaw
with GPT-5.5 has a comparable runtime (293.42~s), but obtains a slightly lower
score (0.7797). OpenCode with GPT-5.5 uses the fewest total tokens (273,532 per
task), yet its Final score is only 0.6906. These results suggest that the
choice of an autonomous agent should account for the desired trade-off between
quality, speed, and computational cost.

\subsection{RQ4. What Are the Common Failure Modes of Autonomous Agents on \ourbench?}
\label{sec:failure-analysis}

To characterize residual failures, we analyze DuMate, the best-performing
agent on \ourbench, and Claude Code, a widely used production agent. For each of DuMate and Claude Code, we stratify the non-complete RQ1 runs
by the four evaluated base LLMs, GPT-5.5, Opus-4.8, GLM-5.2, and
DeepSeek-V4-Pro, and randomly sample 50 runs in total while preserving the
model strata. For each run, we inspect the execution trace
and evaluator feedback and assign one primary failure category.
Table~\ref{tab:failure-categories} reports the category distributions, while
Table~\ref{tab:dumate-case-studies} illustrates how three common failures arise
in concrete DuMate workflows.

\begin{table}[t]
    \centering
    \caption{Primary failure categories among 50 analyzed non-complete runs
    for DuMate and Claude Code.
    Percentages are computed within each agent's sample.}
    \label{tab:failure-categories}
    \scriptsize
    \setlength{\tabcolsep}{3pt}
    \renewcommand{\arraystretch}{1.10}
    \begin{tabularx}{\columnwidth}{
        >{\raggedright\arraybackslash}X
        >{\centering\arraybackslash}p{0.19\columnwidth}
        >{\centering\arraybackslash}p{0.19\columnwidth}}
        \toprule
        \textbf{Failure category} &
        \textbf{DuMate} &
        \textbf{Claude Code} \\
        \midrule
        Incomplete execution or budget exhaustion
        & 14 (28\%) & 16 (32\%) \\
        Incorrect implementation or tool use
        & 15 (30\%) & 11 (22\%) \\
        Requirement or context grounding failure
        & 7 (14\%) & 18 (36\%) \\
        Environment or dependency failure
        & 12 (24\%) & 3 (6\%) \\
        Other failures
        & 2 (4\%) & 2 (4\%) \\
        \bottomrule
    \end{tabularx}
\end{table}

Among the sampled non-complete runs, two failure patterns recur for both
agents in Table~\ref{tab:failure-categories}. Incomplete execution or budget exhaustion accounts for 28\% of the
DuMate sample and 32\% of the Claude Code sample. Incorrect implementation or
tool use accounts for a further 30\% and 22\%, respectively. These results
show that agents often either fail to close the workflow within the required time limit or deliver artifacts
that do not satisfy the task requirements. In addition, requirement or
context grounding failure is more frequent for Claude Code (36\% versus 14\%),
whereas environment or dependency failure is more frequent for DuMate (24\%
versus 6\%). These patterns suggest different improvement priorities: Claude Code would
benefit from stronger context filtering, workspace localization, and
requirement tracking, whereas DuMate would benefit from more robust fallback and recovery strategies for environment and dependency failures.


\begin{table}[t]
    \caption{Representative DuMate failure cases. Each case identifies the
    task, decisive agent error, and consequence.}
    \label{tab:dumate-case-studies}
    \centering
    \scriptsize
    \setlength{\tabcolsep}{3pt}
    \renewcommand{\arraystretch}{1.12}
    \begin{tabularx}{\columnwidth}{
        >{\raggedright\arraybackslash}p{0.25\columnwidth}
        >{\raggedright\arraybackslash}X}
        \toprule
        \textbf{Failure category} &
        \textbf{Task, failure point, and consequence} \\
        \midrule

        Incomplete execution or budget exhaustion &
        \textbf{Task.} Create 20 SVG slides and an inspection report at the
        specified output paths.

        \textbf{Failure point.} Although the task required multiple artifacts
        at exact output paths, the agent spent its remaining budget on
        pixel-level overflow checks after generating the slides, rather than
        exporting the deliverables.

        \textbf{Consequence.} The slides and inspection report were not
        exported before the run reached the time limit of 1800 seconds. \\
        \addlinespace

        Environment or dependency failure &
        \textbf{Task.} Produce a Markdown report on Huaneng Power
        International (600011.SH), including market data, news, sentiment, and
        analysis from external sources.

        \textbf{Failure point.} Under unstable network conditions, web
        search returned a JSON parsing error and the fallback service failed DNS resolution. The agent attempted additional public sources but could not obtain a verified alternative before finalizing the report.

        \textbf{Consequence.} The delivered report marked the required news
        and sentiment sections as ``unverifiable.'' \\
        \addlinespace
        Incorrect implementation or tool use &
        \textbf{Task.} Create an investment-facing report from supplied Word
        documents while keeping every figure consistent with the input content.

        \textbf{Failure point.} Although the task required consistency across
        source documents and the new artifact, the agent inserted an
        unverified share-lot value rather than tracing it to the references.

        \textbf{Consequence.} The report stated 2,000 shares per trading lot,
        whereas the reference specified 2,500. \\
        
        \bottomrule
    \end{tabularx}
\end{table}

We present three cases in Table~\ref{tab:dumate-case-studies} illustrate how the
challenges in \ourbench cause concrete agent failures. In the first
case, DuMate generates all requested slides but spends its remaining time budget on
fine-grained validation and terminates before exporting the required artifacts.
In the second case, successive service failures test the agent's recovery
capability. Despite trying several alternative sources, DuMate cannot obtain
verified evidence and leaves required sections incomplete. In the third case,
DuMate fails to trace a numeric claim to the supplied Word document and inserts a
value that conflicts with the source documents. Together, these cases show how \ourbench exposes weaknesses in the execution
planning, failure recovery, and artifact verification required to complete
complex real-world workflows.

\section{Conclusion}
\label{sec:conclusion}

We introduced \ourbench, a real-session benchmark for evaluating autonomous
agents on complex workflows that require coordinated use of multiple
productivity tools. Its 200 de-identified and human-reviewed tasks are derived
from multi-turn DuMate sessions and reconstruct the user-visible context and
workspace state available when each request was issued. The benchmark combines
compositional tasks with standardized Docker environments that instantiate
insufficient, unstable, and noisy conditions. It further evaluates
heterogeneous artifacts through deterministic checklists and
artifact-specific LLM-as-Judge rubrics, capturing both explicit task completion
and output quality. Experiments with five representative agent frameworks and
four state-of-the-art base models, together with robustness, efficiency, and
other diagnostic analyses, reveal substantial room for improvement in strict
task completion. The results also demonstrate that performance under complex
environmental conditions depends on both the base model and the surrounding
agent framework. We hope \ourbench supports the development of autonomous
agents that can reliably execute compositional workflows beyond clean,
task-ready environments.

\section{Ethical Considerations}
\label{sec:ethical-considerations}

\ourbench is constructed from anonymized and privacy-screened DuMate sessions.
Personally identifiable information, credentials, access tokens, private
endpoints, and other sensitive data are removed, and tasks that cannot be
safely de-identified are excluded. The remaining tasks undergo human review
before inclusion. Because the benchmark reflects usage patterns from a single
platform, it may not represent all users, occupations, or workflows. We
mitigate security and safety risks through isolated containers, restricted
permissions, and task-level workspace boundaries.


\bibliographystyle{ACM-Reference-Format}
\bibliography{sample-base}

\newpage
\clearpage
\appendix



\end{document}